\documentclass{article}
\usepackage{spconf,amsmath,epsfig}

\let\OLDthebibliography\thebibliography
\renewcommand\thebibliography[1]{
  \OLDthebibliography{#1}
  \setlength{\parskip}{0pt}
  \setlength{\itemsep}{0pt plus 0.3ex}
}

\usepackage{graphicx}
\usepackage{latexsym}
\usepackage{times}
\usepackage{soul}
\usepackage{url}
\usepackage[hidelinks]{hyperref}
\usepackage[utf8]{inputenc}
\usepackage[small]{caption}
\usepackage{hyperref}
\usepackage{graphicx}
\usepackage{amsmath}
\usepackage{amsthm}
\usepackage{booktabs}
\usepackage{amssymb}
\usepackage[switch]{lineno}
\usepackage{bbm}
\usepackage{cleveref}
\usepackage{bm}
\usepackage{multirow}
\usepackage[normalem]{ulem}
\usepackage{stfloats}
\usepackage{subcaption}

\usepackage{color}
\usepackage{stfloats}
\usepackage {arydshln}
\usepackage{color}
\graphicspath{{img/}}
\usepackage{algpseudocode}
\usepackage{algorithm}
\algrenewcommand\algorithmicrequire{\textbf{Input:}}
\algrenewcommand\algorithmicensure{\textbf{Output:}}
\begin{document}\sloppy
\title{Federated Skewed Label Learning with Logits Fusion}
\name{Yuwei Wang$^1$,Runhan Li$^{1,2}$,Hao Tan$^{1,2}$,Xuefeng Jiang$^{1,2}$,Sheng Sun$^1$\thanks{Corresponding Author: Sheng Sun. Email: sunsheng@ict.ac.cn.},Min Liu$^{1,3}$,Bo Gao$^4$,Zhiyuan Wu$^{1,2}$}
\address{$^1$Institute of Computing Technology, Chinese Academy of Sciences \\ $^2$University of Chinese Academy of Sciences \space $^3$Zhongguancun Laboratory\\$^4$Beijing Jiaotong University}


\maketitle
\begin{abstract}
Federated learning (FL) aims to collaboratively train a shared model across multiple clients without transmitting their local data. 
Data heterogeneity is a critical challenge in realistic FL settings, as it causes significant performance deterioration due to discrepancies in optimization among local models.
In this work, we focus on label distribution skew, a common scenario in data heterogeneity, where the data label categories are imbalanced on each client. To address this issue, we propose FedBalance, which corrects the optimization bias among local models by calibrating their logits. Specifically, we introduce an extra private weak learner on the client side, which forms an ensemble model with the local model. By fusing the logits of the two models, the private weak learner can capture the variance of different data, regardless of their category. Therefore, the optimization direction of local models can be improved by increasing the penalty for misclassifying minority classes and reducing the attention to majority classes, resulting in a better global model. Extensive experiments show that our method can gain 13\% higher average accuracy compared with state-of-the-art methods.
\end{abstract}

\section{Introduction}
Federated Learning (FL) is a distributed machine learning paradigm that explores collaboration among diverse clients to train models without exchanging their local data. Prevailing FL methods \cite{fedavg,fedprox,scaffold} employ a central server to aggregate clients’ learned local models and construct an updated global model in each communication rounds. The global model is then broadcast to all clients for replacing the weights of local models. 
However, FL faces the challenge of heterogeneous data distributions, accompanied by imbalanced label distributions among clients. The discrepancy among the local label distributions among clients leads to divergence among local optimization objectives. As a result, the global model obtained by averaging the local models will deviate from the global optimal solution, ultimately decreasing system performance \cite{neuralcomputing}. Several methods \cite{moon,fedprox,scaffold} align local models with the global model to tackle heterogeneous data among clients while also tolerating imbalanced categories of local data. However, they fail to utilize the statistical information about the data of each category on the client under heterogeneous label distributions. 

To further tackle the issue of label distribution skew prevalent among clients, a series of methods \cite{fedrs,fedrod,fedlc} have been proposed. Notably, FedRoD suggests that a coherent local training objective can be established by setting a common objective for the clients, namely, successfully classifying all classes with their learned local models. Such an objective can be achieved without requiring clients to have knowledge of each other's data. 
Nevertheless, the learning of minority classes is often hindered by the majority classes, where the missing classes can be seen as the extreme case of a few classes \cite{logitadjustment}, it is necessary to balance the local models’ optimization on the majority class data and improve their ability to learn from all categories of data. This motivates us to align the optimization objectives between the local model and the global model in terms of improving the learning ability of the local model for the missing classes and the few classes.

In this work, we propose a novel method to deal with skewed label distributions, named FedBalance, which constructs an integrated model for each client by introducing a weak learner to work with the local model. 
The weak learner is locally trained and the logits generated by it reflect the learning ability of the model fully influenced by the local unbalanced data, The local model, which we consider as the strong learner, is continuously updated via model aggregation and has better performance than the weak one. Therefore, by fusing the logits of the two models, the weak learner will affect the update of the local model by increasing the penalty for the misclassification of minority classes and reducing the attention to the majority classes. Finally, by improving the learning capability of the local model for all classes, a global model with satisfactory performance can be obtained by only aggregating the local models of each client.

In general, the main contributions of this paper are summarized as follows:
\begin{itemize}
    \item 
    We propose FedBalance to solve the issue of label distribution skew among clients. In FedBalance, a weak learner combined with logits fusion technique is conducted to guide local model updates on each client, reducing the misclassification of minority classes and avoiding overlearning majority classes.
    \item 
    Extensive experiments demonstrate that FedBalance can improve the accuracy of the global model and promote the prediction reliability of local models, and can adapt to a variety of real-world applications.
\end{itemize}

\section{Related Work}
\label{RW}
\subsection{Federated Learning over Heterogeneous Data}
Existing methods that improve the performance of FL over heterogeneous data are twofolds: optimizing global aggregation strategies, and optimizing local training strategies. From the perspective of optimizing global aggregation, Li et al. theoretically study the convergence of FedAvg under Non-IID data \cite{convergencefedavg}. Instead of simply averaging the weights of local models, FedMA \cite{fedma} uses a non-parametric Bayesian approach that aggregates the parameters at the layer level. FedNova \cite{fednova} adaptively adjusts the aggregation weights to eliminate objective inconsistency by normalizing local gradients before averaging. FedAdp \cite{fedadp} assigns weights to each model by calculating the two-by-two similarity of the gradients uploaded by each client. FedOpt \cite{fedopt} suggests the application of federated versions of adaptive optimizers, such as ADAGRAD, ADAM, and YOGI. From the perspective of optimizing local training, \cite{moon,fedprox,scaffold} try to design a variety of loss functions to regularize their update direction, which mitigates inter-model bias by limiting local model updates. Specifically, FedProx \cite{fedprox} adds a proximal term to narrow the distance between local model parameters and global model parameters. SCAFFOLD \cite{scaffold} utilizes control variates containing update orientation information for the respective model to correct the local update. 
In addition, MOON introduces contrastive loss to maximize the consistency between the current features extracted by local models and those extracted by the global model. 

\subsection{Logits Information Fusion in Federated Learning}
Logits information fusion is a common technique in FL, which often relies on knowledge distillation \cite{hinton2015distilling,wu2021spirit} to address various challenges in FL \cite{wu2023survey}, such as data heterogeneity \cite{yao2023fedgkd,lee2022preservation,wu2022exploring}, personalization \cite{wu2023fedict,jin2022personalized}, communication efficiency \cite{wu2023fedcache}, noisy labels \cite{jiang2022towards}, etc. Specifically, \cite{yao2023fedgkd,lee2022preservation} fuses historical and current logits information during local training, preventing the local model from drifting away from the global optimization objective. \cite{wu2023fedict,jin2022personalized} fuses both generalized and personalized logits objectives, and achieves better performance on local data. \cite{jiang2022towards} fuses logits information with different confidence levels to mitigate local models to overfit the local datasets containing noisy labels.
\section{Preliminary and Motivation}
\subsection{Problem Formulation}
In FL, each client, denoted as $M_i$, trains a local model on its dataset $D_i$ consisting of $N_i$ samples. A central server oversees the coordination of model aggregation and communication between clients. The classic global objective is defined as the weighted average of the local objectives, expressed as:

\begin{equation}
  F(w) = \sum_{i=1}^{M} p_{i}f_{i}(w_{i}),
\end{equation}
Here, $p_i = \frac{N_i}{\sum_{j=1}^{M}N_j}$ represents the weighting factor for each client's dataset size, $N_i$.The overarching goal is comprised of the individual objectives of each client, denoted as $f_i(w_i)$, where $w_i$ represents the parameters learned by that specific client. The formula for $f_i(w_i)$ involves averaging the loss function $\mathcal{L}(x_j, y_j, w_i)$ across a subset of the client's local data, denoted as $N_i$. For instance, Cross Entropy loss is often adopted:
\begin{equation}
f_i(w_{i})=\frac{1}{N_{i}}\sum_{j=1}^{N_{i}}\mathcal{L}(x_{j},y_{j},w_{i}),
\end{equation}
During each round, a group of clients are chosen at random and upload their models to the server. These models have been improved via multiple local training epochs of local training. Then, the server consolidates the individual updates into one global model. Subsequently, the server disseminates the updated global model to the individual clients for further local training.

\subsection{Non-IID Setting}
Suppose the local data of each client $D_{i}$ obeys the distribution: $\mathcal{P}_i(x,y)=\mathcal{P}_{i}(x|y)\mathcal{P}_{i}(y)$. Due to the imbalanced label distribution, each client may have missing classes, minority classes and majority classes, and $\mathcal{P}_{i}(y)$ differs across each client. So local models based on discrepant data varied greatly among clients.
Moreover, in order to improve the classification accuracy, the local model classifies the minority class data into the majority class as well with a great probability.
This greatly inhibits the ability of local models to learn the minority class.
Therefore, it is hard to get a high-quality global model by aggregating local models.

\subsection{Insight Formulation}
According to the above discussion, one way to mitigate the influence of label imbalance is to make local models classify all classes well. This way can reduce the inter-model bias while improving the learning ability of local models.

In previous works for tackling label distribution skew, i.e. FedLC \cite{fedlc}, FedRoD \cite{fedrod} and FedRS \cite{fedrs}, they perform the same degree of deflation for the same category of logit, where FedLC leverages the number of each category and FedRS introduces the same hyperparameters for the same class. However, they only consider the differences among classes and do not take into account the specificity among data within classes.
The local model also has different learning abilities for different data in the same class, which is reflected in the confidence level of correct classification for each data.
For example, when the two data of majority classes are correctly classified, the data with a higher logit should receive less attention than the other one.
Inspired by this intuition, our work will calibrate the logit according to both data characteristics  and the difference among classes to improve the overall classification capability of local models.

\begin{figure}[t]
\centering
\includegraphics[width=1.0\linewidth]{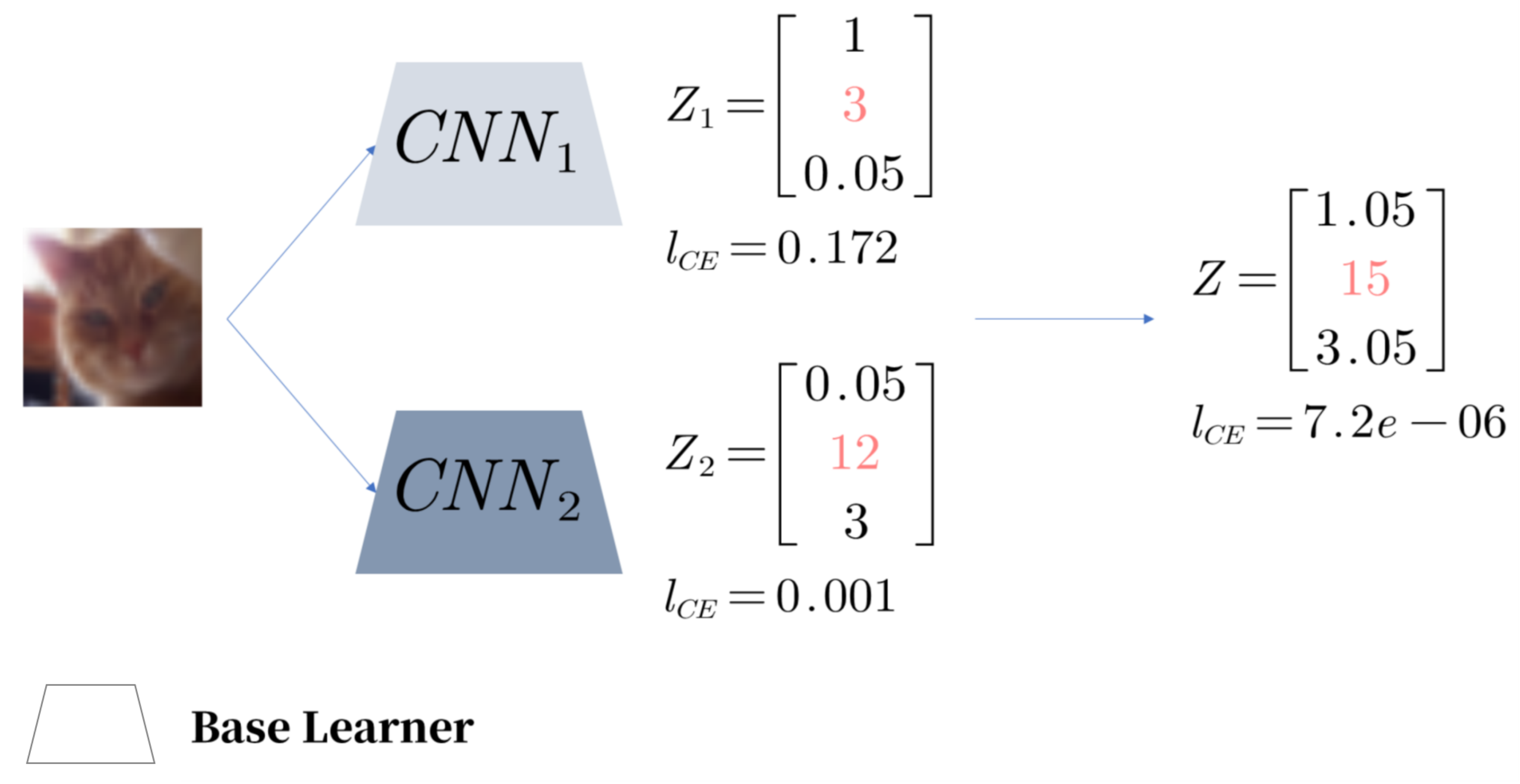}
\caption{Logits and corresponding cross-entropy loss for a ensemble model with two base heterogeneous learners for multiclass scenario. The output corresponding to the true class label is denoted in red font. The darker the color of learner, the higher the certainty of the classification.}
\label{logits}
\end{figure}

\section{Method}

\subsection{Logits Fusion}
Logits Fusion is a decision fusion strategy in Ensemble Learning, aiming to aggregate logit of all base learners in a Ensemble Model:

$$
F_{fusion}(x) = \sum_{j = 1}^{M} F_{j}(x),
$$
where $F_{j}(x)$ is the logit of $j$-th base learner for $x$. However, when ensemble model consists of  heterogeneous models, all learners exhibit a different kinds of certainty for the same input. According to \cite{confidence}, the larger the model capacity, the easier it will be to make the cross-entropy loss smaller and become confident while achieving lower classification error during the training process. Further, as shown in Fig. \ref{logits}, the final prediction of ensemble model is dominated by the most confident learner \cite{average} and if the learners classify correctly, it gets a smaller cross-entropy loss. In other words, if all the learners classify correctly, logits fusion approach weakens the magnitude of contribution to confident learners.

It inspires us to set a smaller capacity model as the private weak learner $\psi$, making the optimization objective of the ensemble model a surrogate for the optimization objective of the local model $\phi$. Meanwhile, by fusing logit before softmax, the attention of the local model $\phi$ on the easily classified samples will be weakened.

\begin{figure}[t]
    \centering
  \includegraphics[width=1.0\linewidth]{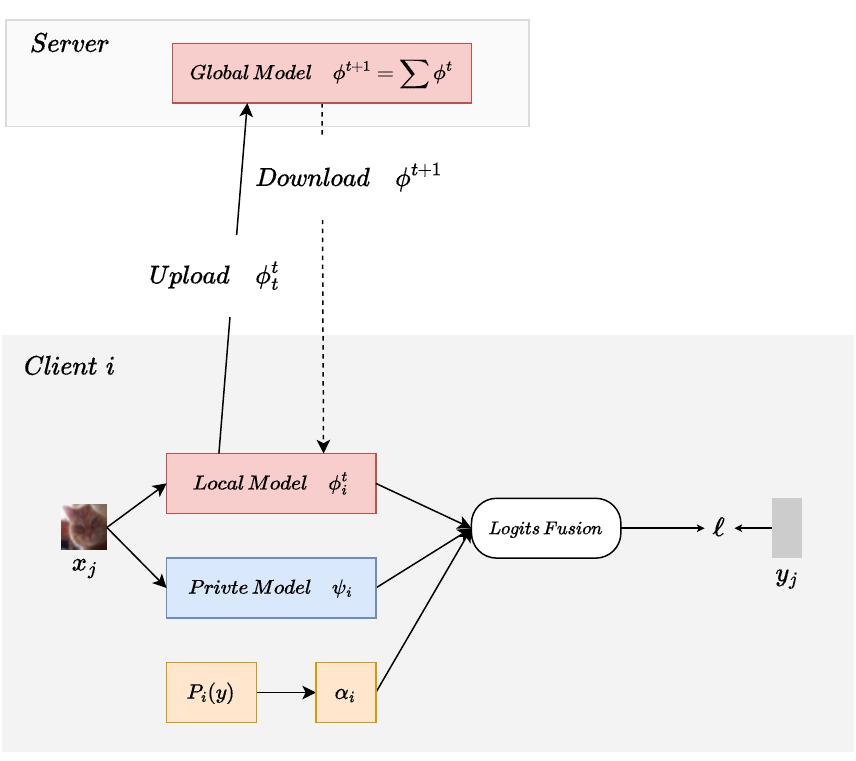}
  \caption{The framework of our method. $P_i(y)$ is label distribution of $i$-th client data.}
  \label{framework}
\end{figure}

\subsection{Weight Fusion}
When facing label skew data, we find that the under-optimized learner $\psi$ only can capture the feature of majority class samples. For majority classes, the attention of the local model $\phi$ on samples belong to these classes will be weakened. On the contrary, for minority classes, $\psi$ will offer a very small value to encourage local model $\phi$ to claim a larger logit. The poor classification performance of the privacy model for minority classes will exacerbate the penalty of the local model for these classes. Using category information to distinguish minority classes from majority classes, we choose $\alpha$ as fusion weight:

$$
\alpha \,\,=\,\,\left\{ \frac{n_i}{N} \right\} |_{i=0}^{C},
$$
where $C$ is the number of class.

\subsection{Local Objective}
Our local objective for $i$-th client is as follows:
\begin{equation}
  \label{equ:loss}
  \begin{split}
    f_i(w_{i}^{\phi})=\mathbb{E}_{(x,y)\sim D_i}[\mathcal{L}(\phi_i(x)+\alpha_i \psi_i(x)|w_{i}^{\phi},w_{i}^{\psi})].
  \end{split}
\end{equation}
As shown in Fig. \ref{framework}, we set a private weak learner $\psi$ on each client aiming to calibrate the logit of local model $\phi$. The privacy model $\psi$ is only trained on the client, thus it actually represents a biased model that is fully influenced by label imbalance data in real life. This means that the privacy model $\psi$ is difficult to capture feature representations of the minority classes but has relatively better classification results for the majority classes. While the local model is continuously updated via model aggregation. For simplicity, we describe our method in Algorithm \ref{alg:wordy}.

\begin{algorithm}[t!]
  \caption{FedBalance}\label{alg:wordy}
  \begin{algorithmic}[1]
    \Require{communication rounds $R$, local training epochs $E$, number of all clients $M$, fraction of clients to sample $\gamma$}
    \Ensure{The final global model $\phi^{R}$}
    \State Initialize $\phi^{1}$ and $\psi_0^1,...,\psi_M^1$ for all clients
    \For{$r=1,2,...,R$}
    \State $S^r \gets $ random sample $\gamma \times M$ clients
    \For{$i=1,2,\dots,\gamma \times M$} in parallel
    \State $\phi_{i}^{r+1}, \psi^{r+1}_i \gets$ \Call{LocalTrain}{$\phi^{r}, \psi^{r}_i$}
    \EndFor
    \State $\phi^{r+1}\gets \frac{1}{n_{S^r}}\sum_{i\in S^r}{n_i}\phi_{i}^{r+1}$
    \EndFor
    \State \textbf{Return} $\phi^{R}$
    \Statex
    \Function{LocalTrain}{$\phi^{r},\psi_i^r$}
    \For{$k=1,2,\dots,K$}
    \State Compute $\mathcal{L}$ by Eq. \ref{equ:loss}
    \State Update $\phi_i^{r},\psi_i^r$ by backpropagation
    \EndFor
    \State \textbf{Return} $\phi^{r+1}_i,\psi^{r+1}_i $
    \EndFunction
  \end{algorithmic}
\end{algorithm}

\section{Experiment}

\subsection{Experimental Setup}
\subsubsection{Datasets and Non-IID Data Partition}
We conduct a number of experiments on popular image classification benchmark datasets: CIFAR-10, CIFAR-100 \cite{cifar} and CINIC-10 \cite{cinic},. Furthermore, to prove the effectiveness of our method for a more practical setting, we experiment on a medical dataset COVID-19 Radiography \cite{covid1,covid2}, which contains chest X-ray images for COVID-19 positive cases as well as Normal and Viral Pneumonia images, including 3616 COVID-19 positive cases along with 10,192 Normal, 6012 Lung Opacity (Non-COVID lung infection) and 1345 Viral Pneumonia images. For image preprocessing, all images are resized to $32\times32$ for all datasets.

The Dirichlet distribution is mostly used to model real-world clients' data distribution, therefore we use it to perform label imbalance-based data partitioning \cite{dirichlet}.
The training data drawn from the Dirichlet distribution with a concentration parameter $\beta$ is assigned to each client.
Specifically, we sample $p_k \sim Dir_N(\beta)$ and allocate the $p_{k,j}$ proportion of the instances of class $k$ to client $j$, where $Dir(\cdot )$ denotes the Dirichlet distribution and $\beta$ is a concentration parameter ($\beta > 0$). The smaller $\beta$ is, the more unbalanced the partitioning is. For ease of presentation, we denote the partition strategy by $p_k \sim Dir_N(\beta)$, which holds for each client $j$.

\subsubsection{Baselines and Implementation Details}
All experiments were conducted using the PyTorch framework \cite{pytorch}, while utilizing the GeForce RTX 3090 GPU. To ensure reproducibility, we anonymously uploaded the code to \href{https://anonymous.4open.science/r/FedBalance_ECML}{Anonymous GitHub}. We use the SGD optimizer with a fixed learning rate of 0.01 in our experiments, along with L2 weight decay and momentum set to 1e-5 and 0.9, respectively. The batch size is fixed at 64, and the number of local epochs defaults to 10. The number of communication rounds is defined as 200 for COVID-19 and 500 for other datasets. We use ResNet-32 \cite{resnet} for CIFAR-100, and ResNet-8 for other datasets as the base model structure. Additionally, As our base learners, we utilize Alexnet\cite{alexnet} and LeNet\cite{lenet}, respectively. Since the aim of our methodology is to alleviate the impact of data heterogeneity and enhance the generalization performance of the global model, we select related state-of-the-art approaches that focus on Non-IID issues as our baselines. Details are provided below:

\begin{itemize}

    \item \textbf{Aggregation-optimized Approaches:} We implemented FedAvg, FedNova and FedOpt, which are detailed discussed in section \ref{RW}. As they did not make any direct changes to the local model, the experiments related to the local model will not involve those methods.
    \item \textbf{Local Training Optimization Approaches:} We implemented FedProx, MOON, FedRS and FedRod, which are detailed discussed in section \ref{RW}. For FedProx, $\mu=0.5$ is adopted as the default value and the results verify the validity of the parameter. For MOON, we leverage two 128-dim Linear Layers as projection layers and set the default hyperparameter $\mu=0.5$ consistent with the original paper. For FedRS, we set $\alpha=0.5$, the best-performing parameter value in the original paper, as the default value. For FedRod, we set $\mu=0.5$ as the default value. Our method FedBalance does not require any additional hyperparameters to be set up.
    
\end{itemize}

\subsubsection{Evaluating local model accuracy across all categories} We conducted a statistical analysis on the average accuracy rates of all local clients' categories, referred to as \textbf{Per-Class Acc} in the CIFAR-10 dataset, while considering a non-i.i.d. distribution with $\beta=0.1$. To accomplish this, we utilized a total of 20 clients, with 4 participating in each aggregation round.Upon completion of each round, we calculated the mean accuracy rates of all categories for the models trained by the 4 clients and computed their average value.

\subsection{Comparison with State-of-the-art Methods}

\begin{table*}[thbp]
  \setlength{\tabcolsep}{4pt}
  \renewcommand{\arraystretch}{1.4}
  \resizebox{\textwidth}{!}{
    \begin{tabular}{llllllllll}
      \hline
      \multirow{2}{*}{\textbf{Methods}} & \multicolumn{3}{|c|}{\textbf{CIFAR-10}}               & \multicolumn{3}{c|}{\textbf{CINIC-10}} & \multicolumn{3}{c}{\textbf{CIFAR-100}}                                                                                                                                                                                                                                                                                                                                                           \\
      \cline{2-10}
                                        & \multicolumn{1}{|c}{$\beta=0.1$}                      & \multicolumn{1}{c}{$\beta=0.3$}        & \multicolumn{1}{c|}{$\beta=0.5$}                      & \multicolumn{1}{|c}{$\beta=0.1$}                    & \multicolumn{1}{c}{$\beta=0.3$}                     & \multicolumn{1}{c|}{$\beta=0.5$}                      & \multicolumn{1}{|c}{$\beta=0.1$}                       & \multicolumn{1}{c}{$\beta=0.3$}                      & \multicolumn{1}{c}{$\beta=0.5$}                      \\
      \hline
      FedAvg                            & \multicolumn{1}{|c}{45.55\small{$\pm$7.22}}           & 65.61\small{$\pm$2.93}                 & \multicolumn{1}{c|}{68.42\small{$\pm$2.81}}           & \multicolumn{1}{c}{34.14\small{$\pm$13.33}}         & \multicolumn{1}{c}{44.64\small{$\pm$3.03}}          & \multicolumn{1}{c|}{53.04\small{$\pm$1.72}}           & \multicolumn{1}{|c} {47.20\small{$\pm$0.80}}           & \multicolumn{1}{c} {47.87\small{$\pm$0.60}}          & \multicolumn{1}{c} {48.06\small{$\pm$0.59}}          \\
      FedNova                           & \multicolumn{1}{|c}{43.23\small{$\pm$3.22}}           & 63.09\small{$\pm$0.51}                 & \multicolumn{1}{c|}{64.91\small{$\pm$0.60}}           & \multicolumn{1}{c}{37.44\small{$\pm$2.08}}          & \multicolumn{1}{c}{45.91\small{$\pm$1.55}}          & \multicolumn{1}{c|}{51.79\small{$\pm$1.13}}           & \multicolumn{1}{|c}  {38.55\small{$\pm$0.32}}          & \multicolumn{1}{c} {43.63\small{$\pm$0.41}}          & \multicolumn{1}{c} {44.53\small{$\pm$0.29}}          \\
      FedOpt                            & \multicolumn{1}{|c}{44.42\small{$\pm$9.41}}           & 66.99\small{$\pm$4.16}                 & \multicolumn{1}{c|}{68.65\small{$\pm$4.17}}           & \multicolumn{1}{c}{25.84\small{$\pm$7.02}}          & \multicolumn{1}{c}{41.09\small{$\pm$3.34}}          & \multicolumn{1}{c|}{47.69\small{$\pm$3.93}}           & \multicolumn{1}{|c}  {45.24\small{$\pm$0.95}}          & \multicolumn{1}{c} {51.03\small{$\pm$0.65}}          & \multicolumn{1}{c} {\uline{52.46\small{$\pm$0.98}}}  \\
      \hdashline

      FedProx                           & \multicolumn{1}{|c}{48.74\small{$\pm$3.23}}           & 62.77\small{$\pm$1.17}                 & \multicolumn{1}{c|}{64.96\small{$\pm$1.47}}           & \multicolumn{1}{c}{\uline{37.50\small{$\pm$3.89}}}  & \multicolumn{1}{c}{46.45\small{$\pm$2.04}}          & \multicolumn{1}{c|}{52.61\small{$\pm$1.60}}           & \multicolumn{1}{|c}{41.36\small{$\pm$0.58}}            & \multicolumn{1}{c} {39.97\small{$\pm$0.49}}          & \multicolumn{1}{c} {39.68\small{$\pm$0.46}}          \\
      MOON                              & \multicolumn{1}{|c}{43.29\small{$\pm$7.99}}           & 68.73\small{$\pm$3.32}                 & \multicolumn{1}{c|}{\uline{71.07\small{$\pm$2.58}} }  & \multicolumn{1}{c}{24.12\small{$\pm$6.93}}          & \multicolumn{1}{c}{44.88\small{$\pm$2.81}}          & \multicolumn{1}{c|}{52.40\small{$\pm$1.72}}           & \multicolumn{1}{|c} {44.52\small{$\pm$0.98}}           & \multicolumn{1}{c} {47.63\small{$\pm$0.49}}          & \multicolumn{1}{c} {48.56\small{$\pm$0.53}}          \\

      \hdashline
      FedRS                             & \multicolumn{1}{|c}{\uline{56.94\small{$\pm$2.23}} }  & \uline{68.78\small{$\pm$1.23}}         & \multicolumn{1}{c|}{70.78\small{$\pm$1.71}}           & \multicolumn{1}{c}{36.94\small{$\pm$2.04}}          & \multicolumn{1}{c}{\uline{50.65\small{$\pm$1.45}}}  & \multicolumn{1}{c|}{54.49\small{$\pm$1.16}}           & \multicolumn{1}{|c}  {47.43\small{$\pm$0.36}}          & \multicolumn{1}{c} {48.18\small{$\pm$0.44}}          & \multicolumn{1}{c} {48.58\small{$\pm$0.38}}          \\
      FedRod                            & \multicolumn{1}{|c}{56.17\small{$\pm$2.56}}           & 68.74\small{$\pm$1.35}                 & \multicolumn{1}{c|}{70.92\small{$\pm$1.71}}           & \multicolumn{1}{c}{\uline{37.50\small{$\pm$2.99}}}  & \multicolumn{1}{c}{50.56\small{$\pm$1.52}}          & \multicolumn{1}{c|}{\uline{54.72\small{$\pm$1.28}} }  & \multicolumn{1}{|c}  {\uline{47.52\small{$\pm$0.49}}}  & \multicolumn{1}{c} {\uline{52.33\small{$\pm$0.37}}}  & \multicolumn{1}{c} {48.61\small{$\pm$0.35}}          \\

      \hline
      \textbf{Ours(Lenet)}              & \multicolumn{1}{|c}{66.70\small{$\pm$1.71}}           & \textbf{75.80\small{$\pm$1.10}}        & \multicolumn{1}{c|}{\textbf{77.78\small{$\pm$1.10}} } & \multicolumn{1}{c}{42.88\small{$\pm$1.84}}          & \multicolumn{1}{c}{50.76\small{$\pm$1.57}}          & \multicolumn{1}{c|}{55.51\small{$\pm$0.89} }          & \multicolumn{1}{|c}  {\textbf{49.57\small{$\pm$0.54}}} & \multicolumn{1}{c} {\textbf{54.49\small{$\pm$0.50}}} & \multicolumn{1}{c} {54.11\small{$\pm$0.34}}          \\
      \textbf{Ours(Alexnet)}                     & \multicolumn{1}{|c}{\textbf{67.90\small{$\pm$1.65}} } & 75.71\small{$\pm$0.80}                 & \multicolumn{1}{c|}{77.29\small{$\pm$0.91} }          & \multicolumn{1}{c}{\textbf{43.02\small{$\pm$1.97}}} & \multicolumn{1}{c}{\textbf{51.04\small{$\pm$1.66}}} & \multicolumn{1}{c|}{\textbf{55.81\small{$\pm$1.01}} } & \multicolumn{1}{|c}  {49.46\small{$\pm$0.55}}          & \multicolumn{1}{c} {53.61\small{$\pm$0.44}}          & \multicolumn{1}{c} {\textbf{55.17\small{$\pm$0.51}}} \\
      \hline
    \end{tabular}
  }
  \caption{Test accuracy (\%) for learning with three different degrees of Non-IID on CIFAR-10, CINIC-10, CIFAR-100 and COVID-19. Repeat all experiments three times and report the mean and standard derivation. The bolded numbers represent the best performance, and the underlined numbers represent the optimal performance of the baseline.}
  \label{main_res}
\end{table*}

\begin{figure*}[t]

  \begin{subfigure}{0.33\linewidth}
    \centering
    \includegraphics[width=\linewidth]{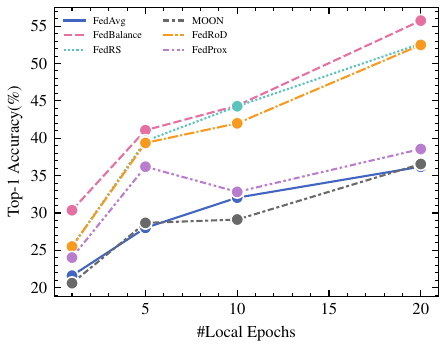}
    \caption{$\beta=0.1$}
  \end{subfigure}
  \begin{subfigure}{0.33\linewidth}
    \centering
    \includegraphics[width=\linewidth]{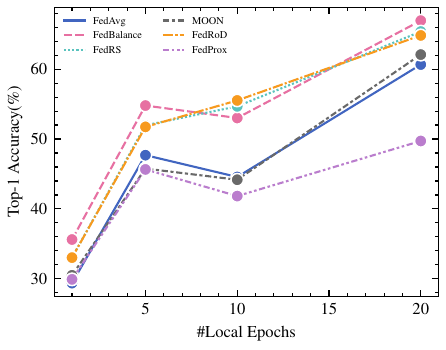}
    \caption{$\beta=0.3$}
  \end{subfigure}
  \begin{subfigure}{0.33\linewidth}
    \centering
    \includegraphics[width=\linewidth]{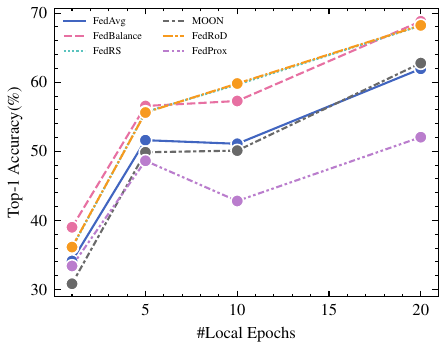}
    \caption{$\beta=0.5$}
  \end{subfigure}

  \caption{Accuracy of the global model with different local epochs for each method on CIFAR10 under different non-independent identically distributed scenarios.}
  \label{acc_peochs}

\end{figure*}

\subsubsection{Main Results and Analysis}
In this section, we present a comparison of the generalization performance of FedBalance against the aforementioned baselines across all datasets, with all results illustrated on Table \ref{main_res}. Our experimental results demonstrate that our method outperforms all baselines across most non-i.i.d. scenarios. It is worth noting that the accuracy of aggregation-based methods is generally lower than ours, as such methods fail to fully consider category information across different clients, particularly when faced with extreme label imbalance distributions. 
On the other hand, local training optimization-based methods, such as FedProx and MOON, strive to obtain a more generalized global model by bringing local models closer towards the global model. However, in some scenarios, these methods perform even worse than vanilla FedAvg. This can be attributed to the underlying assumption that the global model usually demonstrates better generalization ability, but blindly approaching it can lead to a lack of local category information integration during training, ultimately degrading overall performance. While FedRS and FedRod did achieve relatively high accuracy, they did not take into account client-side optimization of majority class samples and were only aimed at specific categories during learning. As a result, their accuracy is still lower than that of our method. It is worth noting that our method yields significant performance improvements even for the most extreme non-i.i.d. data partition scenarios with $\beta=0.1$, leading to an increase in accuracy from $56.94$ (the highest performing baseline) to $67.90$ on CIFAR-10. This strongly indicates the potential and validity of our approach. Furthermore, our method exhibits the smallest standard deviation across most tests, indicating a smaller performance gap among locally trained models. We will provide more detailed information on this in the following section. Lastly, experimental results indicate that our approach is independent of the specific structure of the private weak learner, as we have tested various models with different structures.

\subsubsection{Number of Local Update} We conducted a study on the impact of the number of local training epochs on the final model accuracy, with the results shown in Fig. \ref{acc_peochs}. It is evident that the performance of the global model increasingly improves as the number of client-side local updates increases. Notably, even with a small number of local epochs, our method is still able to perform well. This highlights the efficiency of our approach in learning client-side features. Furthermore, when faced with deep degrees of non-independent distribution, such as $\beta=0.1$, most methods are limited by label imbalance and can no longer achieve significant improvements despite increasing the number of local epochs. In contrast, our approach yields the most outstanding result. This demonstrates the effectiveness of our method in mitigating the negative impact of imbalanced local data on client-side models.

\subsubsection{Accuracy Testing on Medical Dataset} In this section, we evaluate the performance of all methods on the medical COVID-19 Dataset. To better simulate real-world scenarios, we set $\beta=0.3$ and divide the data into 20 clients. In each round, we randomly select 4 clients and present the final accuracy of the global model.  As can be seen from Table 2, our method achieves the highest accuracy in all non-i.i.d. scenarios. This demonstrates the significant importance of our approach in real-world medical applications.

\begin{table}
  \centering
  \renewcommand{\arraystretch}{1.2}
  \begin{tabular}{llll}
    \hline
    \multirow{2}{*}{\textbf{Methods}} & \multicolumn{3}{|c}{\textbf{COVID-19}}                                                                                                       \\ 
    \cline{2-4}
                                      & \multicolumn{1}{|c}{$\beta=0.1$}                     & \multicolumn{1}{c}{$\beta=0.3$} & \multicolumn{1}{c}{$\beta=0.5$}                     \\
    \hline
    FedAvg                            & \multicolumn{1}{|c}{32.10\small{$\pm$10.14}}         & 60.06\small{$\pm$14.01}         & \multicolumn{1}{c}{71.07\small{$\pm$8.58}}          \\
    FedOpt                            & \multicolumn{1}{|c}{35.21\small{$\pm$14.10}}         & 40.27\small{$\pm$13.13}         & \multicolumn{1}{c}{53.96\small{$\pm$13.84}}         \\
    FedNova                           & \multicolumn{1}{|c}{\uline{40.62\small{$\pm$9.44}}}  & 56.37\small{$\pm$6.79}          & \multicolumn{1}{c}{62.35\small{$\pm$5.18}}          \\
      \hdashline
    FedProx                           & \multicolumn{1}{|c}{37.38\small{$\pm$11.93}}         & 58.42\small{$\pm$8.82}          & \multicolumn{1}{c}{71.75\small{$\pm$4.57}}          \\
    MOON                              & \multicolumn{1}{|c}{29.72\small{$\pm$11.06}}         & 56.53\small{$\pm$14.08}         & \multicolumn{1}{c}{69.10\small{$\pm$8.08}}          \\
      \hdashline
    FedRS                             & \multicolumn{1}{|c}{56.59\small{$\pm$4.85}}          & 69.39\small{$\pm$8.05}          & \multicolumn{1}{c}{72.73\small{$\pm$9.83}}          \\
    FedRod                            & \multicolumn{1}{|c}{\uline{57.09\small{$\pm$5.07}}}  & \uline{70.28\small{$\pm$7.58}}  & \multicolumn{1}{c}{\uline{72.90\small{$\pm$9.94}}}  \\

    \hline
    \textbf{Ours}                     & \multicolumn{1}{|c}{\textbf{58.20\small{$\pm$4.52}}} & \textbf{71.30\small{$\pm$3.83}} & \multicolumn{1}{c}{\textbf{75.74\small{$\pm$3.25}}} \\
    \hline
  \end{tabular}
  \caption{Examining Test Accuracy (\%) on COVID-19 Dataset with Three Degrees of Non-IID Learning. Please conduct all experiments in triplicate and report the mean and standard deviation. The bold numbers indicate the optimal performance, while the underlined numbers indicate the baseline's optimal performance.}
  \label{main_res}
\end{table}




\begin{figure*}[t]

  \begin{subfigure}{0.33\linewidth}
    \centering
    \includegraphics[width=\linewidth]{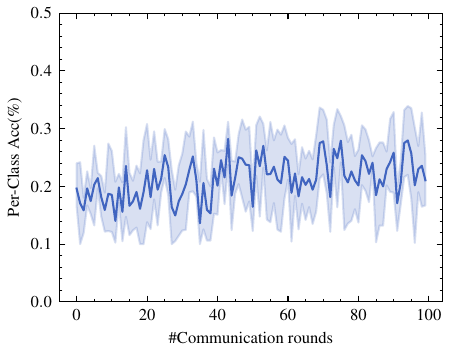}
    \caption{FedAvg}
  \end{subfigure}
  \begin{subfigure}{0.33\linewidth}
    \centering
    \includegraphics[width=\linewidth]{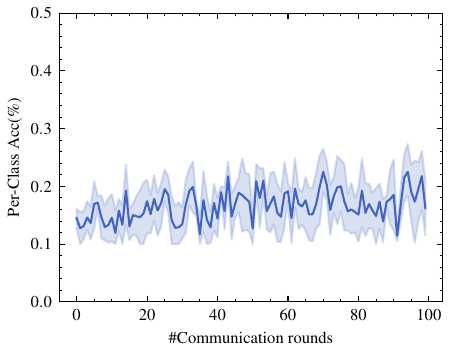}
    \caption{FedProx}
  \end{subfigure}
  \begin{subfigure}{0.33\linewidth}
    \centering
    \includegraphics[width=\linewidth]{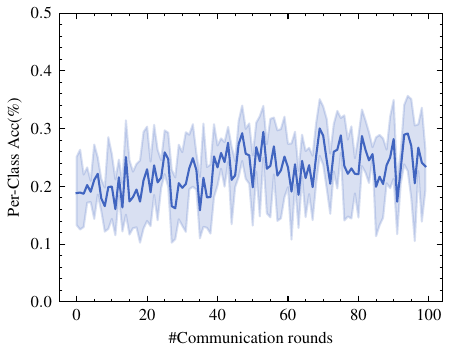}
    \caption{MOON}
  \end{subfigure}\\
  \begin{subfigure}{0.33\linewidth}
    \centering
    \includegraphics[width=\linewidth]{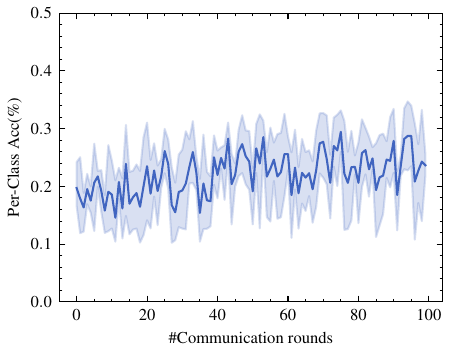}
    \caption{FedRS}
  \end{subfigure}
  \begin{subfigure}{0.33\linewidth}
    \centering
    \includegraphics[width=\linewidth]{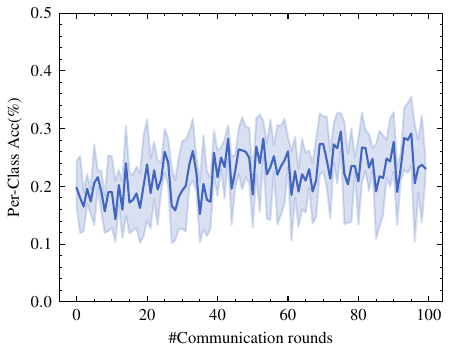}
    \caption{FedRoD}
  \end{subfigure}
  \begin{subfigure}{0.33\linewidth}
    \centering
    \includegraphics[width=\linewidth]{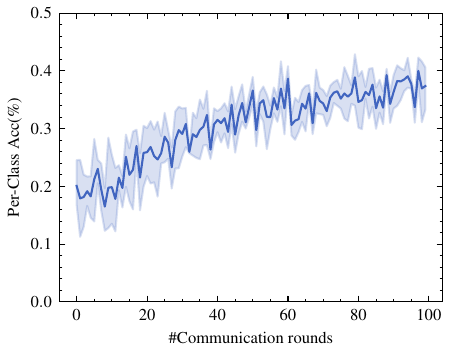}
    \caption{Our Method}
  \end{subfigure}
  \caption{Mean Per-Class Accuracy of 4 Randomly Selected Clients on CIFAR-10 with $\beta$=0.1 for 100 rounds. The shaded area in the graph represents the 95\% confidence interval for the per-class accuracies of clients, whereas the solid line corresponds to the mean value.}
  \label{Per-class_all}

\end{figure*}


\subsubsection{Local Model Variance}
\label{localmodel}
In this study, we investigate the extent of deviation among all client models during the federated training process, as shown in Fig. \ref{Per-class_all}. Based on the figure, it can be observed that our approach results in the smallest shaded area, indicating a certain degree of alignment among the optimization goals among clients. Notably, in our approach, the Per-class Accuracy of clients continued to improve as the number of rounds increased, despite the existence of missing and minority classes in each client's local data. These findings suggest that our approach leads to improved generalization performance of the local models trained.

\subsubsection{Communication Efficiency and Computational Overhead}
 Our findings, reported in Table \ref{overheads}, indicate the number of communication rounds required by our approach to achieve the same level of accuracy as FedAvg. Remarkably, our approach achieves a speedup that is nearly 3.33 times that of FedAvg, clearly demonstrating that our approach greatly enhances communication efficiency. We also quantified the local model computational overhead associated with each method. Given that a significant proportion of the computation in deep neural networks is typically dominated by matrix multiplication operations, which essentially equate to MAC (Multiply-Accumulate Operations) operations, the number of MAC operations can provide a reliable estimate of the overall computational complexity of a model. This makes MAC operations an effective metric for comparing the computational costs of diverse models, and for optimizing their architecture to reduce computational complexity.  Specifically, when using LeNet as the private weak learner, our computational overhead increased by only 0.05\% compared to FedAvg. These results suggest that our approach results in only minimal computational overhead while offering higher accuracy and more efficient communication.

 \begin{table}
  \centering
  \renewcommand{\arraystretch}{1.4}
  \begin{tabular}{|c|c|c|c|}
    \hline
    \multicolumn{1}{l|}{\textbf{Methods}} & \multicolumn{1}{c|}{\textbf{Round}} & \multicolumn{1}{c}{\textbf{SpeedUp}} & \multicolumn{1}{|c}{\textbf{MACs(M)}}     \\

    \hline
    \multicolumn{1}{l|}{FedAvg}           & \multicolumn{1}{c|}{100}            & \multicolumn{1}{c}{1.00$\times$}       & \multicolumn{1}{|c}{12.75}          \\
    \hline
    \multicolumn{1}{l|}{FedProx}          & \multicolumn{1}{c|}{98}             & \multicolumn{1}{c}{1.05$\times$}       & \multicolumn{1}{|c}{12.75}          \\
    \hline
    \multicolumn{1}{l|}{MOON}             & \multicolumn{1}{c|}{95}             & \multicolumn{1}{c}{1.05$\times$}       & \multicolumn{1}{|c}{38.25} \\
    \hline
    \multicolumn{1}{l|}{FedRS}            & \multicolumn{1}{c|}{50}             & \multicolumn{1}{c}{2.00$\times$}       & \multicolumn{1}{|c}{12.75} \\
    \hline
    \multicolumn{1}{l|}{FedRod}           & \multicolumn{1}{c|}{50}             & \multicolumn{1}{c}{2.00$\times$}       & \multicolumn{1}{|c}{12.75} \\
    \hline
    \multicolumn{1}{l|}{Ours(Lenet)}      & \multicolumn{1}{c|}{30}             & \multicolumn{1}{c}{\textbf{3.33$\times$}}       & \multicolumn{1}{|c}{13.41} \\
    \hline
    \multicolumn{1}{l|}{Ours(Alexnet)}    & \multicolumn{1}{c|}{30}             & \multicolumn{1}{c}{\textbf{3.33$\times$}}       & \multicolumn{1}{|c}{27.74} \\
    \hline
  \end{tabular}

  \caption{Comparing Round Numbers to Achieve Same Accuracy as 100 Rounds of FedAvg on CIFAR-10 with $\beta=0.1$ and Computational Overhead. The acceleration of each method is computed with reference to FedAvg.}
  \label{overheads}
\end{table}

\subsection{Ablation Study}

\begin{table}
  \centering
    \renewcommand{\arraystretch}{1.4}
    \begin{tabular}{|ccc|}
      \hline
      \multicolumn{1}{l|}{\textbf{Methods}}              & \multicolumn{1}{c}{\textbf{Accuracy}}         & \multicolumn{1}{|c}{\textbf{KL}}       \\

      \hline
      \multicolumn{1}{l|}{ResNet-8+ResNet-8}             & \multicolumn{1}{l}{26.31{$\pm$1.34}}          & \multicolumn{1}{|l}{3.03e-03}          \\
      \hline
      \multicolumn{1}{l|}{ResNet-8+Alexnet (Softmax)}      & \multicolumn{1}{l}{24.98{$\pm$5.26}}          & \multicolumn{1}{|l}{1.65e-02}          \\
      \hline
      \multicolumn{1}{l|}{ResNet-8+Alexnet (Ours)}       & \multicolumn{1}{l}{\textbf{46.26{$\pm$3.10}}} & \multicolumn{1}{|l}{\textbf{1.95e-05}} \\
      \hline
    \end{tabular}

  \caption{Analyzing Global Model Accuracy and KL Divergence for Local and Integrated Model Outputs on a Randomly Selected Client During Local Training. }
  \label{Ablation}
\end{table}

In this study, we investigate the impact of different degrees of local model dominance in the integrated model on the performance of the global model. The results of the experiments are presented in Table \ref{Ablation}. When the private weak learner has the same structure as the local model, a higher KL dispersion indicates that the output of the integrated model deviates significantly from the output of the local model. This is because the output of the integrated model cannot be dominated by the local model, which hinders the learning of the local model, resulting in a decrease in accuracy. The rationale behind the poor performance of probability fusion is similar. After softmax, the output of both the local and private weak learners are scaled to the range of 0 to 1, which also cannot allow the output of the integrated model to be dominated by the local model.

\section{Conclusion}
We propose FedBalance, a federated learning method for local data distributions with skewed labels. FedBalance uses a private weak learner and a logits fusion technique to guide the local model optimization on each client, so as to reduce the misclassification of minority classes and avoid overfitting to majority classes. Extensive experiments on both simulation and real-world datasets demonstrate the effectiveness of FedBalance, with higher average accuracy compared with state-of-art methods.

\bibliographystyle{IEEEbib}

\end{document}